\pdfoutput=1

\documentclass[11pt]{article}

\usepackage{EACL2023}

\usepackage{times}
\usepackage{latexsym}

\usepackage[T1]{fontenc}

\usepackage[utf8]{inputenc}

\usepackage{microtype}

\usepackage{inconsolata}

\usepackage{bm}
\usepackage{amsmath,amssymb}
\usepackage{bbm}
\usepackage{multirow}
\usepackage{makecell}
\usepackage{pifont}
\usepackage{subfigure}
\usepackage{arydshln}
\usepackage{graphicx}

\newcommand{\setR}{\mathcal{R}}
\newcommand{\setX}{\mathcal{X}}
\newcommand{\setE}{\mathcal{E}}
\newcommand{\setM}{\mathcal{M}}
\newcommand{\setV}{\mathcal{V}}
\newcommand{\setT}{\mathcal{T}}

\title{DREEAM: Guiding Attention with Evidence \\for Improving Document-Level Relation Extraction}

\author{Youmi Ma \and An Wang \and Naoaki Okazaki \\
Tokyo Institute of Technology \\
\texttt{\{youmi.ma@nlp., an.wang@nlp., okazaki@\}c.titech.ac.jp} \\
}

\begin{document}
\maketitle
\begin{abstract}

Document-level relation extraction (DocRE) is the task of identifying all relations between each entity pair in a document.
Evidence, defined as sentences containing clues for the relationship between an entity pair, has been shown to help DocRE systems focus on relevant texts, thus improving relation extraction.
However, evidence retrieval (ER) in DocRE faces two major issues: high memory consumption and limited availability of annotations.
This work aims at addressing these issues to improve the usage of ER in DocRE.
First, we propose DREEAM, a memory-efficient approach that adopts evidence information as the supervisory signal, thereby guiding the attention modules of the DocRE system to assign high weights to evidence.
Second, we propose a self-training strategy for DREEAM to learn ER from automatically-generated evidence on massive data without evidence annotations.
Experimental results reveal that our approach exhibits state-of-the-art performance on the DocRED benchmark for both DocRE and ER.
To the best of our knowledge, DREEAM is the first approach to employ ER self-training\footnote{The source code is available at \url{https://github.com/YoumiMa/dreeam}}.

\end{abstract}

\section{Introduction}

\begin{figure}[t]
    \centering
    \includegraphics[width=.46\textwidth]{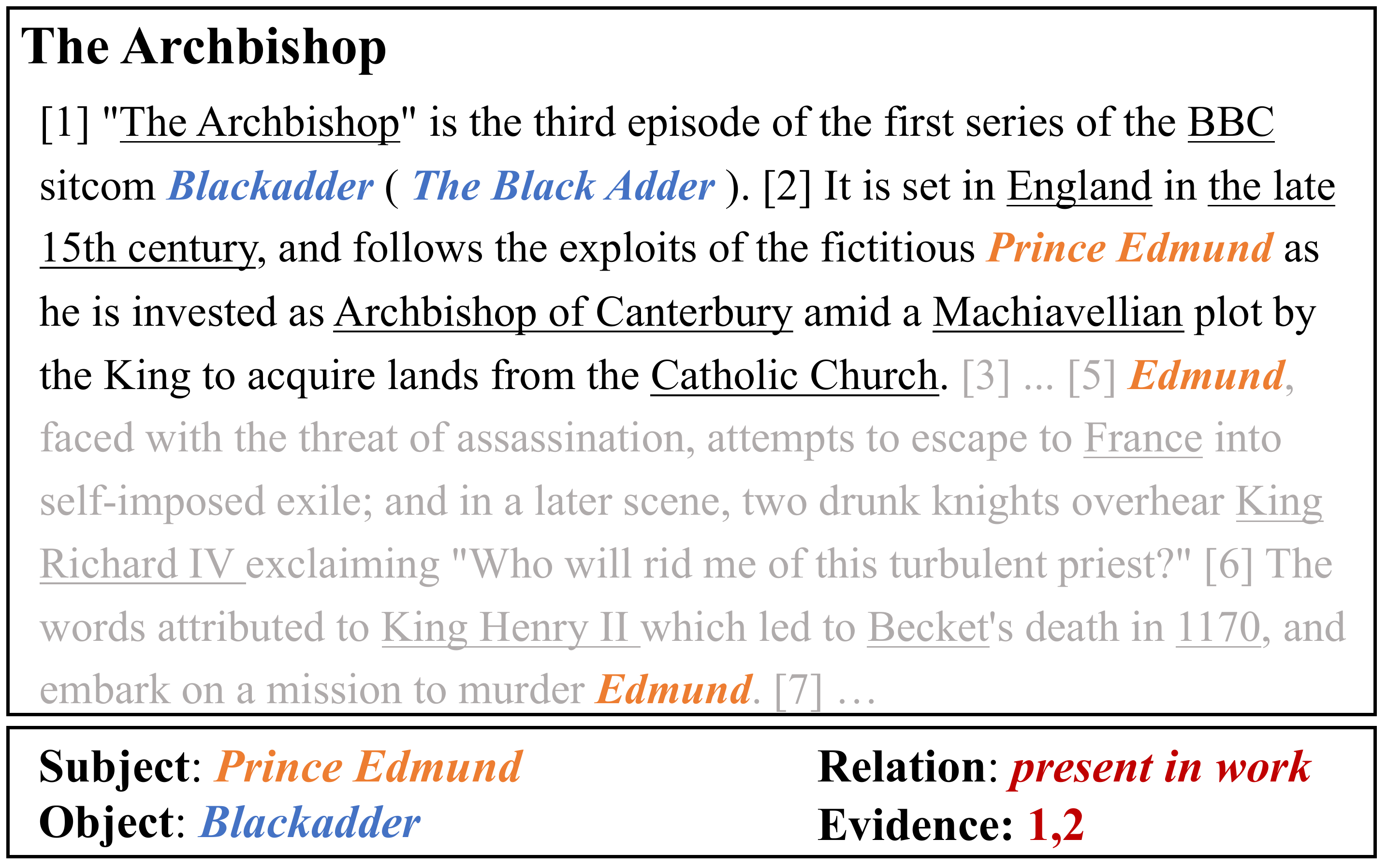}
    \caption{Example document and one of the relation triples from DocRED, where the $i$-th sentence is marked with [i] in the beginning. Mentions in bold italics are those of subjects and objects, whereas entity mentions other than subject and object are underlined.}
    \label{fig:example}
\end{figure}

Document-level relation extraction (DocRE) has been recognized as a more realistic and challenging task compared with its sentence-level counterpart~\cite{peng-etal-2017-cross,verga-etal-2018-simultaneously,yao-etal-2019-docred}.
In DocRE, an entity can have multiple mentions scattered throughout a document, and relationships can exist between entities in different sentences.
Therefore, DocRE models are expected to apply information filtering to long texts by focusing more on sentences relevant to the current decision of relation extraction (RE) and less on irrelevant ones.
To this end, existing studies retrieve \textit{supporting evidence} (evidence hereafter, \citealp{yao-etal-2019-docred}), a set of sentences necessary for humans to identify the relation between an entity pair~\cite{huang-etal-2021-entity,huang-etal-2021-three,xie-etal-2022-eider,xiao-etal-2022-sais,xu-etal-2022-document}.
As shown in Figure~\ref{fig:example}, to decide the \textit{present in work} relation between \textit{Prince Edmund} and \textit{Blackadder}, reading sentences 1 and 2 should be sufficient.
Although sentences 5 and 6 also mention the subject, they are irrelevant to the relation decision.
Evidence of the relation triple (\textit{Prince Edmund}, \textit{present in work}, \textit{Blackadder}) is thus defined as sentences 1 and 2.

Despite the usefulness of evidence, automatic evidence retrieval (ER) faces two major issues.
Firstly, the existing approaches for ER are memory-inefficient.
Previous systems tackle ER and DocRE as separate tasks, introducing extra neural network layers to learn ER with DocRE jointly~\cite{huang-etal-2021-entity,xie-etal-2022-eider,xiao-etal-2022-sais}.
The ER module typically involves a bilinear classifier that receives entity-pair-specific embeddings and sentence embeddings as the input.
To compute the evidence score of each sentence for each entity pair, the module must walk through all (entity pair, sentence) combinations.
The computations significantly increase memory consumption, particularly in documents with numerous sentences and entities.
Secondly, the availability of human annotations of evidence is limited.
To make matters worse, gold training data for DocRE are more expensive to annotate than those for their sentence-level counterpart.
Despite the difficulty of obtaining human annotations, acquiring evidence annotations at a low cost has been underexplored.
Although automatically collecting silver training data for RE by distant supervision~\cite{mintz-etal-2009-distant,yao-etal-2019-docred}, locating evidence for a sliver RE instance in the document is nontrivial.

This work aims at alleviating these issues to improve the usage of ER in DocRE.
To reduce the memory consumption, we propose \textbf{D}ocument-level \textbf{R}elation \textbf{E}xtraction with \textbf{E}vidence-guided \textbf{A}ttention \textbf{M}echanism (DREEAM), a memory-efficient approach for incorporating DocRE with ER.
We adopt ATLOP~\cite{zhou2021atlop}, a Transformer-based DocRE system widely used in previous studies~\cite{xie-etal-2022-eider,tan-etal-2022-document,xiao-etal-2022-sais}, as the backbone.
Instead of introducing an external ER module, we directly guide the DocRE system to focus on evidence.
Specifically, we supervise the computation of entity-pair-specific local context embeddings.
The local context embedding, formed as a weighted sum among all token embeddings based on attention from the encoder, is trained to assign higher weights to evidence and lower weights otherwise.

To compensate for the shortage of evidence annotations, we propose performing ER under a weakly-supervised setting.
Specifically, we design a strategy to perform self-training with DREEAM on massive, unlabeled data.
The data is obtained from distant supervision (distantly-supervised data hereafter) and thus is automatically annotated with relation labels but not evidence labels.
We expect the knowledge about ER learned from the human-annotated data to generate and grow on the distantly-supervised data.
To enable self-training, we first adopt a teacher model trained on human-annotated data to retrieve silver evidence from distantly-supervised data.
Next, we train a student model on the data for RE while learning ER from the silver evidence.
The student model is further finetuned on the human-annotated data to refine its knowledge.
Experiments on the DocRED benchmark~\cite{yao-etal-2019-docred} show that with the help of ER self-training, DREEAM exhibits state-of-the-art performance on both RE and ER.

In short, the contributions of this work are: (1) We propose DREEAM, a memory-efficient approach to incorporate evidence information into Transformer-based DocRE systems by directly guiding the attention.
DREEAM does not introduce any extra trainable parameters for ER while achieving good performance on both RE and ER.
(2) We propose incorporating distantly-supervised RE training with ER self-training, which improves the performance on both tasks. 
To the best of our knowledge, DREEAM is the first DocRE system that enables joint training of ER and RE under a weakly-supervised setting.

\section{Preliminary}
\subsection{Problem Formulation}

Given a document $D$ containing sentences $\setX_D = \{x_i\}_{i=1}^{|\setX_D|}$ and entities $\setE_D = \{e_i\}_{i=1}^{|\setE_D|}$, DocRE aims to predict all possible relations between every entity pair. 
Each entity $e\in \setE_D$ is mentioned at least once in $D$, with all its proper-noun mentions denoted as $\setM_e=\{m_i\}_{i=1}^{|\mathcal{M}_e|}$.
Each entity pair $(e_s, e_o)$ can hold multiple relations, comprising a set $\setR_{s,o} \subset \setR$, where $\setR$ is a pre-defined relation set.
We let the set $\setR$ include $\epsilon$, which stands for \textit{no-relation}.
Additionally, if an entity pair $(e_s, e_o)$ carries a valid relation $r\in \setR \backslash \{\epsilon\}$, ER aims to retrieve the supporting evidence $\setV_{s,r,o} \subseteq \setX_D$ that are sufficient to predict the triplet $(e_s, r, e_o)$.

\subsection{ATLOP}
\label{sec:model}
This section reviews ATLOP, the backbone of our proposed method.
\paragraph{Text Encoding} Before encoding, a special token \texttt{*} is inserted at the beginning and the end of each entity mention.
Then, tokens $\setT_D=\{t_i\}_{i=1}^{|\setT_D|}$ within document $D$ are encoded with a Transformer-based pretrained language model (PLM, \citealp{NIPS2017attention}) to obtain token embeddings and cross-token dependencies.
Although the original ATLOP adopts only the last layer, this work takes the average of the last three layers\footnote{Pilot experiments showed that using the last 3 layers yields better performance than using only the last layer.}.
Specifically, for a PLM with $d$ hidden dimensions at each transformer layer, the token embeddings $\bm{H}$ and cross-token dependencies $\bm{A}$ are computed as:
\begin{equation}
    \label{eq:text_enc}
    \bm{H}, \bm{A} = \textrm{PLM}(\setT_D), 
\end{equation}
where $\bm{H} \in \mathbb{R}^{|\setT_D| \times d}$ averages over hidden states of each token from the last three layers and $\bm{A} \in \mathbb{R}^{|\setT_D| \times |\setT_D|}$ averages over attention weights of all attention heads from the last three layers.
\paragraph{Entity Embedding}  The entity embedding $\bm{h}_{e} \in \mathbb{R}^d$ for each entity $e$ with mentions $\setM_{e}=\{m_i\}_{i=1}^{|\mathcal{M}_e|}$ is computed by collecting information from all its mentions. 
Specifically, \texttt{logsumexp} pooling, which has been empirically shown to be effective in previous studies~\cite{jia-etal-2019-document}, is adopted as: $\bm{h}_{e} = \log\sum_{i=1}^{|\setM_e|}\exp(\bm{H}_{m_i})$, where $\bm{H}_{m_i}$ is the embedding of the special token \texttt{*} at the starting position of mention $m_i$.

\paragraph{Localized Context Embedding} To better utilize information from long texts, ATLOP introduces entity-pair specified localized context embeddings. 
Intuitively, for entity pair $(e_s,e_o)$, tokens important to both $e_s$ and $e_o$ should contribute more to the embedding.
The importance of each token is determined by the cross-token dependencies $\bm{A}$ obtained from Equation~\ref{eq:text_enc}.
For entity $e_s$, the importance of each token is computed using the cross-token dependencies of all its mentions $\setM_{e_s}$. 
First, ATLOP collects and averages over the attention $\bm{A}_{m_i} \in \mathbb{R}^{|\setT_D|}$ at the special token \texttt{*} before each mention $m_i \in \setM_{e_s}$ to get $\bm{a}_s \in \mathbb{R}^{|\setT_D|}$ as the importance of each token for entity $e_s$.
Then, the importance of each token for an entity pair $(e_s,e_o)$, noted as $\bm{q}^{(s,o)} \in \mathbb{R}^{|\setT_D|}$, is computed from $\bm{a}_s$ and $\bm{a}_o$ as: 
\begin{equation}
\bm{q}^{(s,o)} = \frac{\bm{a}_s \circ \bm{a}_o}{\bm{a}_s^\top \bm{a}_o}, \label{eq:context_emb_weight} 
\end{equation}
where $\circ$ stands for the Hadamard product.
$\bm{q}^{(s,o)}$ is thus a distribution that reveals the importance of each token for entity pair $(e_s, e_o)$.
Subsequently, ATLOP performs a localized context pooling,
\begin{equation}
\bm{c}^{(s,o)} = \bm{H}^\top \bm{q}^{(s,o)},\label{eq:context_rep}
\end{equation}   
where $\bm{c}^{(s,o)}\in \mathbb{R}^d$ is a weighted average over all token embeddings. 
\paragraph{Relation Classification} To predict the relation between entity pair $(e_s, e_o)$, ATLOP first generates context-aware subject and object representations:
\begin{align}
    & \bm{z}_s = \tanh(\bm{W}_s[\bm{h}_{e_s};\bm{c}^{(s,o)}] + \bm{b}_s) \\
    & \bm{z}_o = \tanh(\bm{W}_o[\bm{h}_{e_o};\bm{c}^{(s,o)}] + \bm{b}_o), 
\end{align}
where $[\cdot;\cdot]$ represents the concatenation of two vectors and $\bm{W}_s, \bm{W}_o \in \mathbb{R}^{d \times 2d}, \bm{b}_s, \bm{b}_o \in \mathbb{R}^d$ are trainable parameters. 
Then, a bilinear classifier\footnote{In practice, a grouped bilinear classifier~\cite{Zheng2019LearningDB} is applied to save memory.} is applied on the context-aware representations to compute the relation scores $\bm{y}^{(s,o)} \in \mathbb{R}^{|\setR|}$:
\begin{equation}
    \bm{y}^{(s,o)} = \bm{z}_s^\top\mathsf{W}_r\bm{z}_o + \bm{b}_r,
\end{equation}
where $\mathsf{W}_r \in \mathbb{R}^{|\mathcal{R}| \times d\times d}$ and $\bm{b}_r \in \mathbb{R}^{|\mathcal{R}|}$ are trainable parameters.
The probability that relation $r\in \setR$ holds between entity $e_s$ and $e_o$ is thus $\mathrm{P}(r|s,o) = \sigma(y^{(s,o)}_r)$, where $\sigma$ is the sigmoid function.

\paragraph{Loss Function} ATLOP proposes Adaptive Thresholding Loss (ATL) that learns a dummy threshold class TH during training, serving as a dynamic threshold for each relation class $r\in\setR$.
For each entity pair $(e_s, e_o)$, ATL forces the model to yield scores above TH for positive relation classes $\setR_P$ and scores below TH for negative relation classes $\setR_N$, formulated as below:
\begin{equation}
    \label{eq:atl}
    \begin{split}
    \mathcal{L}_{\mathrm{RE}} = & - \sum_{s \neq o}\sum_{r \in \setR_P} \frac{\exp(y^{(s,o)}_r)}{\sum_{r' \in \setR_P \cup \{\mathrm{TH}\}}\exp(y_{r'}^{(s,o)})}\\
    & - \frac{\exp(y^{(s,o)}_{\mathrm{TH}})}{\sum_{r' \in \setR_N \cup \{\mathrm{TH}\} }\exp(y_{r'}^{(s,o)})}.
    \end{split}
\end{equation}

The idea of setting a threshold class is similar to the Flexible Threshold~\cite{chen-etal-2020-hierarchical}.

\section{Proposed Method: DREEAM}

\begin{figure*}
    \centering
    \subfigure[Model architecture of DREEAM.]{\includegraphics[width=.47\textwidth]{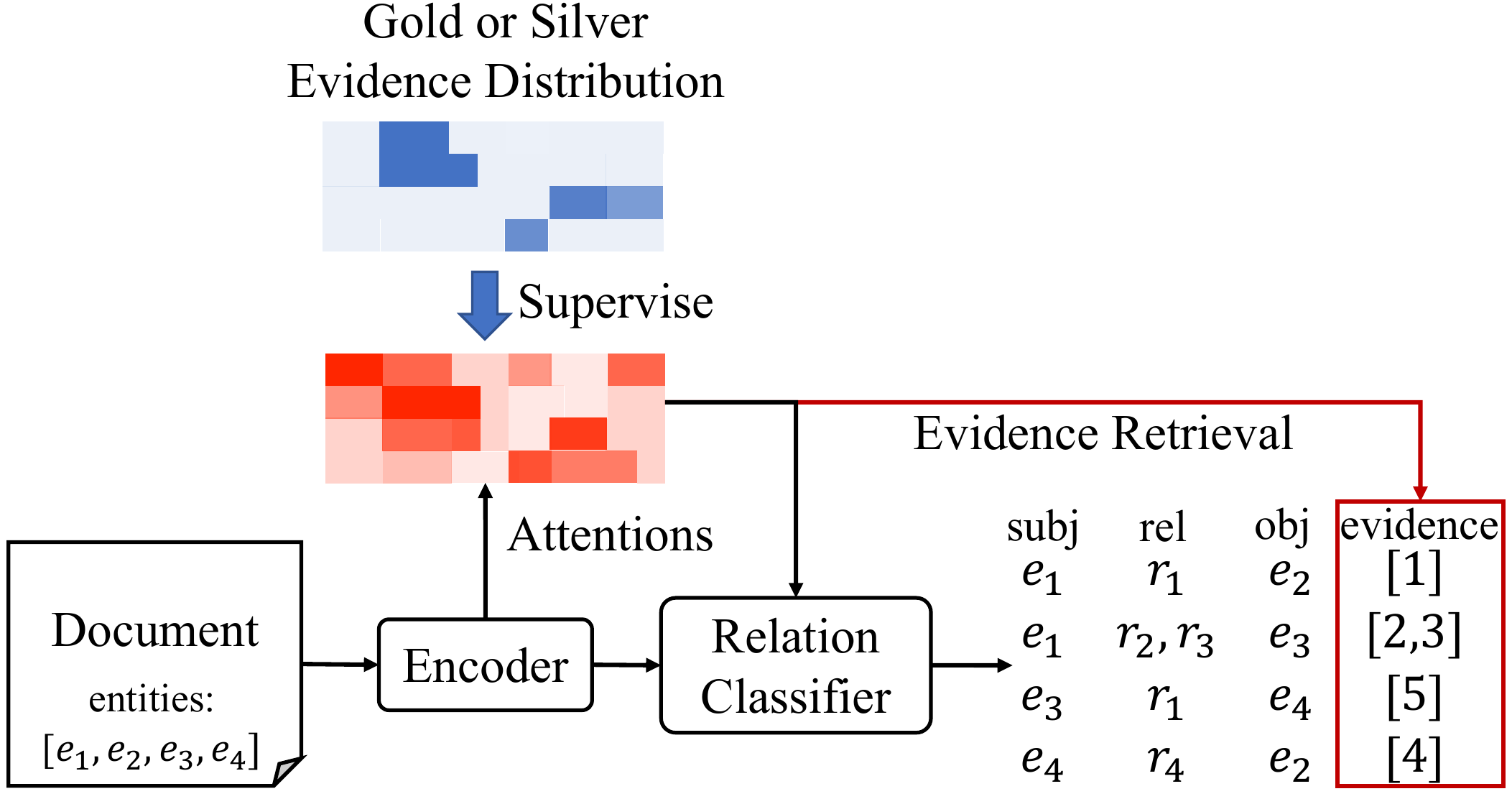}}
    \hspace{.03\textwidth}
    \subfigure[Information flow of self-training using DREEAM.]{\includegraphics[width=.47\textwidth]{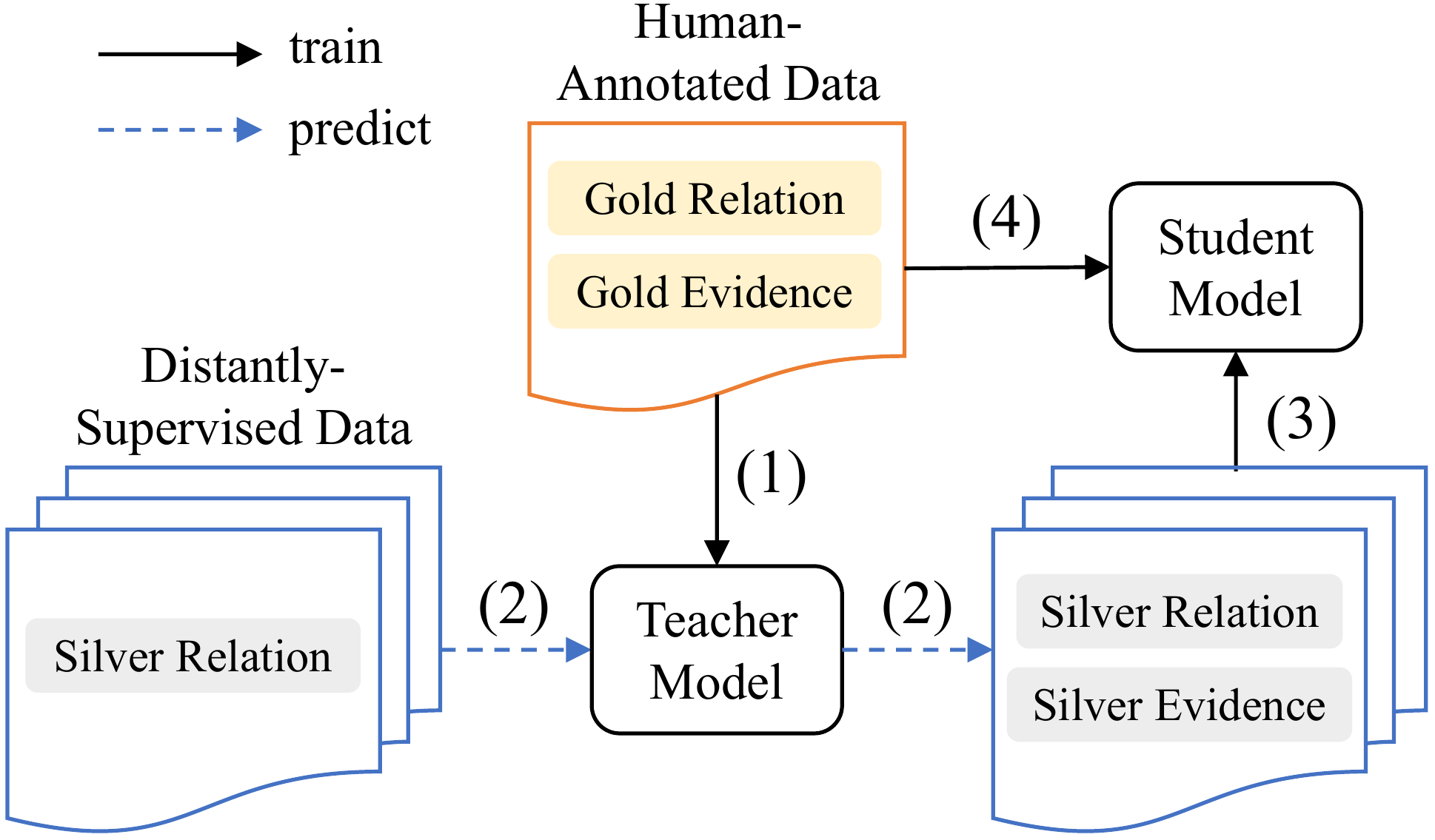}}
    \caption{Model architecture and the information flow during self-training. In (a), gold/silver evidence distributions come from human-annotations/the teacher model. In (b), arrows represent the direction of knowledge transfer.}
    \label{fig:model}
\end{figure*}
To perform information filtering, ATLOP computes a localized context embedding based on attention weights from the Transformer-based encoder.
The rationale is that cross-token dependencies are encoded as attention weights in Transformer layers.
In this work, we propose DREEAM to enhance ATLOP with evidence.
In addition to the automatically-learned cross-token dependencies, the attention modules are supervised to concentrate more on evidence sentences and less on others.

DREEAM can be employed for both supervised and self-training, sharing the same architecture with different supervisory signals, as shown in Figure~\ref{fig:model} (a).
Inspired by~\citet{tan-etal-2022-document}, we propose a pipeline to enable self-training of ER, with the data flow shown in Figure~\ref{fig:model} (b).
First, we train a teacher model on human-annotated data with gold relations and evidence labels.
Next, we apply the trained teacher model to predict silver evidence for the distantly-supervised data.
Then, we train a student model on the distantly-supervised data, with ER supervised by the silver evidence. 
Finally, we finetune the student model on the human-annotated data to refine its knowledge.
The rest of this section introduces the training processes of the teacher and student models, followed by the inference process.

\subsection{Teacher Model}
\label{sec:teacher}
For each entity pair $(s,o)$, we guide $\bm{q}^{(s,o)}$ with an evidence distribution to help generate an evidence-centered localized context embedding.
While $\bm{q}^{(s,o)}$ yields token-level importance for $e_s$ and $e_o$, we can obtain only sentence-level evidence from human annotations, as shown in Figure~\ref{fig:example}.
To alleviate this gap, we sum the weight of each token within a sentence.
Specifically, for a sentence $x_i \in \setX_D$ consisting of tokens $t_{\textrm{START}(x_i)}, \dots, t_{\textrm{END}(x_i)}$, we obtain the sentence-level importance as:
\begin{equation}
    p_i^{(s,o)} = \sum_{j=\textrm{START}(x_i)}^{\textrm{END}(x_i)}q_j^{(s,o)}.
    \label{eq:sent_imp}
\end{equation}
Collecting the importance of all sentences yields a distribution $\bm{p}^{(s,o)} \in \mathbb{R}^{|\setX_D|}$ that expresses the importance of each sentence within the document. 

We further supervise $\bm{p}^{(s,o)}$ for each entity pair $(e_s, e_o)$ using a human-annotated evidence distribution computed from gold evidence.
First, we define a binary vector $\bm{v}^{(s,r,o)} \in \mathbb{R}^{|\setX_D|}$ for each valid relation label $r \in \setR_{s,o} \subset \setR \backslash \{\epsilon\}$ that records whether each sentence $x_i \in \setX_D$ is evidence of the relation triple $(e_s, r, e_o)$ or not.
For example, if $x_i$ is evidence of $(e_s, r, e_o)$, then $v^{(s,r,o)}_i$ is set to 1, and otherwise 0.

Next, we marginalize all valid relations and normalize the marginalized vector to obtain $\bm{v}^{(s,o)}$:
\begin{equation}
    \bm{v}^{(s,o)} = \frac{\sum_{r\in \setR_{s,o}}\bm{v}^{(s,r,o)}}{\sum_{r\in\setR_{s,o}}{\bm{1}^\top \bm{v}^{(s,r,o)}}},
    \label{eq:evi_human}
\end{equation}
where $\bm{1} = (1,1,\dots,1) \in \mathbb{R}^{|\setX_D|}$ is an all-ones vector.
The rationale behind Equation~\ref{eq:evi_human} is that modules before the relation classifier are not explicitly aware of specific relation types.
We thus guide attention modules within the encoder to produce relation-agnostic token dependencies.

\paragraph{Loss Function} Our purpose is to guide $\bm{p}^{(s,o)}$ with human evidence $\bm{v}^{(s,o)}$ to generate an evidence-focused localized context embedding $\bm{c}^{(s,o)}$.
To achieve this, we train the model with Kullback-Leibler (KL) Divergence loss, minimizing the statistical distance between $\bm{p}^{(s,o)}$ and $\bm{v}^{(s,o)}$:
\begin{equation}
    \mathcal{L}_{\mathrm{ER}}^{\rm gold} = -D_{\rm KL}( \bm{v}^{(s,o)} || \bm{p}^{(s,o)}).
    \label{eq:loss_evi}
\end{equation}

During training, we balance the effect of ER loss with RE loss using a hyper-parameter $\lambda$:
\begin{equation}
    \mathcal{L}^{\rm gold} = \mathcal{L}_{\mathrm{RE}} + \lambda\mathcal{L}_{\mathrm{ER}}^{\rm gold}.\label{eq:loss_gold}
\end{equation}

\subsection{Student Model}

We employ the system trained on human-annotated data as a teacher model to support ER self-training on massive data. 
The data, obtained from relation distant-supervision~\cite{mintz-etal-2009-distant}, contains noisy labels for RE but no information for ER.
We train a student model on the data.
Supervision of the student model, similar to that of the teacher model, consists of two parts:
an RE binary cross-entropy loss and an ER self-training loss.

In general, predictions from the teacher model are adopted as the supervisory signal for ER training.
First, we let the teacher model infer on the distantly-supervised data, thereby yielding an evidence distribution over tokens $\hat{\bm{q}}^{(s,o)}$ for each entity pair $(e_s,e_o)$.
Next, we train the student model to reproduce $\hat{\bm{q}}^{(s,o)}$ for each entity pair $(e_s,e_o)$.

\paragraph{Loss Function} The objectives of self-training are identical to those of supervised training.
We train ER of the student model using a KL-divergence loss similar to Equation~\ref{eq:loss_evi}:
\begin{equation}
    \mathcal{L}^{\rm silver}_{\mathrm{ER}} = -D_{\rm KL} (\hat{\bm{q}}^{(s,o)} || \bm{q}^{(s,o)}),
    \label{eq:loss_evi_silver}
\end{equation}
where $\bm{q}^{(s,o)}$ is the student model's evidence distribution over tokens regarding entity pair $(e_s,e_o)$, computed from Equation~\ref{eq:context_emb_weight}.

There are two notable differences between $\mathcal{L}^{\rm silver}_{\mathrm{ER}}$ and $\mathcal{L}_{\mathrm{ER}}^\mathrm{gold}$.
Firstly, the supervisory signal of $\mathcal{L}_{\mathrm{ER}}^\mathrm{gold}$ is sentence-level, while that of $\mathcal{L}^{\rm silver}_{\mathrm{ER}}$ is token-level.
The gap results from the availability of token-level evidence distributions.
On human-annotated data, it is untrivial to obtain token-level evidence distributions from sentence-level annotations.
On distantly-supervised data, however, the evidence distribution over tokens can be easily obtained from predictions of the teacher model.
We thus adopt token-level evidence distributions to provide supervision from a micro perspective for ER self-training.
Secondly, $\mathcal{L}_{\mathrm{ER}}^\mathrm{gold}$ is computed only on entity pairs with valid relation(s), while $\mathcal{L}_{\mathrm{ER}}^\mathrm{silver}$ is computed over all entity pairs within the document.
The design choice is based on the low reliability of relation labels on distantly-supervised data.
As these relation labels are collected automatically, it is possible that some of the annotated relations are irrelevant to the document.
Therefore, it is hard to tell which relations are valid and which are not from the automatic annotations.
For this reason, we compute the loss from all entity pairs to prevent missing important instances.

The overall loss is balanced by the same hyper-parameter $\lambda$ in Equation~\ref{eq:loss_gold}:
\begin{equation}
    \mathcal{L}^{\rm silver} = \mathcal{L}_{\mathrm{RE}} + \lambda\mathcal{L}_{\mathrm{ER}}^{\rm silver}.
\end{equation}

After training on the distantly-supervised data, the student model is further finetuned using the human-annotated data to refine its knowledge about DocRE and ER with reliable supervisory signals.
\subsection{Inference}

Following~\citet{zhou2021atlop}, we apply adaptive thresholding to obtain RE predictions, selecting relation classes with scores higher than the threshold class as predictions.
For ER, we apply static thresholding and choose sentences with importance higher than a pre-defined threshold as evidence.

We further incorporate the \textbf{inference-stage fusion} strategy proposed by~\citet{xie-etal-2022-eider}. 
Specifically, for each predicted relation triple $(e_s, r, e_o)$ associated with evidence prediction $\setV_{s,r,o}$, we create a pseudo-document $\hat{D}_{s,r,o}$ by collecting only evidence sentences $x_i \in \setV_{s,r,o}$.
Then, we feed pseudo-documents into the trained model to re-score the relation triples.
To aggregate the predictions from the pseudo-documents and the entire document, we adopt a blending layer that contains only one parameter $\tau$ representing a threshold.
Each triple $(e_s, r, e_o)$ is chosen as the final prediction only if the summation of its scores on the entire document and pseudo-documents is higher than $\tau$. 
We adjust $\tau$ to minimize the binary cross-entropy loss of RE on the development set.
For more details, we refer the readers to the original paper~\cite{xie-etal-2022-eider}.

\section{Experiments}

To evaluate DREEAM, we conduct experiments under supervised and weakly-supervised settings.
\subsection{Setting}

\begin{table}[t]
    \centering
    \small
    \begin{tabular}{l|rr}
    \Xhline{3\arrayrulewidth}
         \multicolumn{1}{c|}{\textbf{Statistics}} & \multicolumn{1}{c}{Human} & \multicolumn{1}{c}{Distant} \\
         \Xhline{2\arrayrulewidth}
         \# of documents & 3,053/998/1,000 & 101,873\\
         \# of relation types & 97 & 97 \\
         Avg. \# of ent. per doc. & 19.5 & 19.3 \\
         Avg. \# of sent. per doc. & 8.0 & 8.1\\
         Avg. \# of ment. per ent. & 1.3 & 1.3 \\
         Avg. \# of rel. per doc. & 12.5 & 14.8 \\
         Avg. \# of evi. per rel. & 1.6 & -\\
    \Xhline{3\arrayrulewidth}
    \end{tabular}
    \caption{Data statistics of DocRED. \textit{Human} stands for human-annotated data and \textit{Distant} stands for distantly-supervised data. The abbreviations \textit{doc.}, \textit{ent.}, \textit{sent.}, \textit{ment.}, \textit{rel.}, and \textit{evi.} stand for document, entity, sentence, mention, relation, and evidence sentences, respectively.}
    \label{tab:dataset}
\end{table}

\begin{table*}[!t]
    \centering
    \small
    \begin{tabular}{lccccccc}
    \Xhline{3\arrayrulewidth}
    & & \multicolumn{3}{c}{Dev} & \multicolumn{3}{c}{Test} \\
    Method & PLM & Ign F1 & F1 & Evi F1 & Ign F1 & F1 & Evi F1 \\
    \Xhline{3\arrayrulewidth}
    \multicolumn{8}{l}{\textbf{(a) without Distantly-Supervised Data}} \\
        SSAN~\cite{xu-etal-2021-ssan} & BERT\textsubscript{base} & 57.03 & 59.19 & -  & 55.84 & 58.16 & -\\
        ATLOP~\cite{zhou2021atlop} & BERT\textsubscript{base}  & 59.22 & 61.09  & -  & 59.31 &  61.30 & -\\
        E2GRE~\cite{huang-etal-2021-entity} & BERT\textsubscript{base} & 55.22 & 58.72 & 47.12 & - & - & - \\
        DocuNet~\cite{zhang-etal-2021-document} & BERT\textsubscript{base} & 59.86 & 61.83 & -  & 59.93 & 61.86 & - \\
        EIDER~\cite{xie-etal-2022-eider} & BERT\textsubscript{base} & 60.51 & 62.48  & 50.71 & 60.42 & 62.47 & 51.27 \\
        SAIS~\cite{xiao-etal-2022-sais} & BERT\textsubscript{base} & 59.98 & 62.96 & 53.70 & 60.96 & 62.77 & 52.88 \\
        \hdashline
        DREEAM (teacher) & \multirow{2}{*}{BERT\textsubscript{base}}  & 59.60\textsubscript{\textpm{0.15}} & 61.42\textsubscript{\textpm{0.15}} & \multirow{2}{*}{52.08\textsubscript{\textpm{0.10}}} & 59.12 & 61.13  & \multirow{2}{*}{51.71} \\
        + Inference-stage Fusion & & 60.51\textsubscript{\textpm{0.06}} & 62.55\textsubscript{\textpm{0.06}} & & 60.03 & 62.49 &  \\
        \hline
        SSAN~\cite{xu-etal-2021-ssan} & RoBERTa\textsubscript{large}  & 60.25 & 62.08 & -  & 59.47 & 61.42 & -\\ATLOP~\cite{zhou2021atlop} & RoBERTa\textsubscript{large} & 61.32 & 63.18 & - & 61.39 & 63.40 & - \\
        DocuNet~\cite{zhang-etal-2021-document} & RoBERTa\textsubscript{large}  & 62.23 & 64.12 & -  & 62.39 & 64.55 & -\\
        EIDER~\cite{xie-etal-2022-eider} & RoBERTa\textsubscript{large}  & 62.34 & 64.27 & 52.54 & 62.85 & 64.79 & 53.01 \\
        SAIS~\cite{xiao-etal-2022-sais}  & RoBERTa\textsubscript{large} & 62.23 & 65.17 & 55.84 & 63.44 & 65.11 & 55.67 \\
        \hdashline
        DREEAM (teacher) & \multirow{2}{*}{RoBERTa\textsubscript{large}} & 61.71\textsubscript{\textpm{0.09}} & 63.49\textsubscript{\textpm{0.10}} & \multirow{2}{*}{54.15\textsubscript{\textpm{0.11}}} & 61.62 & 63.55 & \multirow{2}{*}{54.01} \\
        + Inference-stage Fusion & & 62.29\textsubscript{\textpm{0.23}} & 64.20\textsubscript{\textpm{0.23}} & & 62.12  & 64.27 \\
    \Xhline{2\arrayrulewidth}
    \multicolumn{8}{l}{\textbf{(b) with Distantly-Supervised Data}} \\
        KD-DocRE~\cite{tan-etal-2022-document} & BERT\textsubscript{base} & 63.38 & 64.81  & - & 62.56 & 64.76 & -\\
        \hdashline
        DREEAM (student) & \multirow{2}{*}{BERT\textsubscript{base}} & 63.47\textsubscript{\textpm{0.02}} & 65.30\textsubscript{\textpm{0.03}} & \multirow{2}{*}{\textbf{55.68}\textsubscript{\textpm{0.04}}} & 63.31 & 65.30  & \multirow{2}{*}{\textbf{55.43}}\\
        + Inference-Stage Fusion & & \textbf{63.92}\textsubscript{\textpm{0.02}} & \textbf{65.83}\textsubscript{\textpm{0.04}} &  & \textbf{63.73} & \textbf{65.87} & \\
        \hline
        SSAN~\cite{xu-etal-2021-ssan} & RoBERTa\textsubscript{large}  & 63.76 & 65.69 & -  & 63.78 & 65.92 & - \\
        KD-DocRE~\cite{tan-etal-2022-document} & RoBERTa\textsubscript{large} & 65.27 & 67.12 & - & 65.24 & 67.28 & - \\
        \hdashline
        DREEAM (student) & \multirow{2}{*}{RoBERTa\textsubscript{large}} & 65.24\textsubscript{\textpm{0.07}} & 67.09\textsubscript{\textpm{0.07}} & \multirow{2}{*}{\textbf{57.55}\textsubscript{\textpm{0.07}}} & 65.20 & 67.22 & \multirow{2}{*}{\textbf{57.34}} \\
        + Inference-Stage Fusion & & \textbf{65.52}\textsubscript{\textpm{0.07}} & \textbf{67.41}\textsubscript{\textpm{0.04}} &  & \textbf{65.47} & \textbf{67.53} & \\ 
    \Xhline{3\arrayrulewidth}
    \end{tabular}
    \caption{Evaluation results on development and test sets of DocRED, with best scores \textbf{bolded}. The scores of existing methods are borrowed from corresponding papers. We group the methods first by whether they utilize the distantly-supervised data or not, then by the PLM encoder. }
    \label{tab:main_results}
\end{table*}

\paragraph{Dataset}  We conduct experiments on DocRED~\cite{yao-etal-2019-docred}\footnote{\url{https://github.com/thunlp/DocRED}}, the largest dataset for DocRE.
As shown in Table~\ref{tab:dataset}, DocRED contains a small portion of human-annotated data and a large portion of distantly-supervised data made by aligning Wikipedia articles with the Wikidata knowledge base~\cite{wikidata}.
This work directly adopts the distantly-supervised data provided in DocRED.

\paragraph{Configuration} We implement DREEAM based on Hugging Face's Transformers~\cite{wolf-etal-2020-transformers}.
Following previous work, we evaluate the performance of DREEAM using BERT\textsubscript{base}~\cite{devlin-etal-2019-bert} and RoBERTa\textsubscript{large}~\cite{Liu2019RoBERTaAR} as the PLM encoder.
The parameter for balancing ER loss with RE loss is set to 0.1 for BERT\textsubscript{base} and 0.05 for RoBERTa\textsubscript{large} when training both the teacher and the student model, chosen based on a grid search from $\lambda \in \{0.05, 0.1, 0.2, 0.3\}$.
We train and evaluate DREEAM on a single Tesla V100 16GB GPU when utilizing BERT\textsubscript{base} and on a single NVIDIA A100 40GB GPU when utilizing RoBERTa\textsubscript{large}.
Details about hyper-parameters and running time are provided in Appendix~\ref{sec:hyparam}.

\paragraph{Evaluation} During inference, sentences $x_i$ with $p_i > 0.2$ computed from Equation~\ref{eq:sent_imp} are retrieved as evidence.
For the evaluation, we adopt official evaluation metrics of DocRED~\cite{yao-etal-2019-docred}: Ign F1 and F1 for RE and Evi F1 for ER.
Ign F1 is measured by removing relations present in the annotated training set from the development and test sets.
We train our system five times, initialized with different random seeds, and report the average scores and standard error of these runs.

\subsection{Main Results}
\label{sec:main_results}

Table~\ref{tab:main_results} lists the performance of the proposed and existing methods.
We select the best-performing model on the development set to make predictions on the test set and submit the predictions to CodaLab for evaluation\footnote{\url{https://codalab.lisn.upsaclay.fr/competitions/365}. Submissions under username \texttt{kgmr15} are from this work.}.

\paragraph{Performance of the Student Model} Table~\ref{tab:main_results} shows that the student model outperforms existing systems on RE by utilizing the distantly-supervised data.
In particular, when adopting BERT\textsubscript{base} as the PLM encoder, DREEAM performs better than KD-DocRE~\cite{tan-etal-2022-document}, the previous state-of-the-art system, by 0.6/1.0 points on Ign F1/F1 for the development set.
On the test set, the improvement reaches 1.1 F1 points on both Ign F1 and F1.
Notably, DREEAM utilizing BERT\textsubscript{base} even performs comparably with SSAN utilizing RoBERTa\textsubscript{large} under the weakly-supervised setting~\cite{xu-etal-2021-ssan}.
When adopting RoBERTa\textsubscript{large} as the PLM encoder, the advantage of DREEAM remains on both development and test sets.
These results support our hypothesis that ER self-training improves RE, which has not been demonstrated by any previous work.

\paragraph{Performance of the Teacher Model} The upper half of Table~\ref{tab:main_results} shows that the teacher model trained on human-annotated data exhibits comparable performance to EIDER~\cite{xie-etal-2022-eider} on both RE and ER.
Although there is a performance gap between DREEAM and SAIS, we attribute it to the difference in supervisory signals.
While DREEAM incorporates RE with only relation-agnostic ER, SAIS is trained under three more tasks: coreference resolution, entity typing, and relation-specific ER~\cite{xiao-etal-2022-sais}.
These extra supervisory signals possibly contribute to the high performance of SAIS.
Apart from the performance, our method has a critical advantage over previous ER-incorporated DocRE systems in memory efficiency.
We provide a detailed discussion in Section~\ref{sec:memory}.

\paragraph{Effectiveness of ER Self-Training} Additionally, we observe that the student model leads the existing systems by a large margin on ER.
As the first approach enabling weakly-supervised ER training, DREEAM utilizes considerable amounts of data without evidence annotation via self-training.
The experimental results reveal that DREEAM improves over the state-of-the-art supervised approaches by approximately 2.0 points on Evi F1.
Therefore, we conclude that our approach to ER self-training succeeds in acquiring evidence knowledge from the relation-distantly-supervised data with no evidence annotation. 

\subsection{Ablation Studies}

\begin{table}[t]
    \centering
    \small
    \begin{tabular}{lccc}
    \Xhline{3\arrayrulewidth}
    Setting & Ign F1 & F1 & Evi F1 \\
    \Xhline{3\arrayrulewidth}
    \multicolumn{4}{l}{\textbf{(a) Teacher Model}} \\
    DREEAM & \textbf{59.60}\textsubscript{\textpm{0.15}} & \textbf{61.42}\textsubscript{\textpm{0.15}} & \textbf{52.08}\textsubscript{\textpm{0.10}} \\
        \textit{w/o} ER training & 59.21\textsubscript{\textpm{0.19}} & 61.01\textsubscript{\textpm{0.20}} & 42.79\textsubscript{\textpm{1.65}}\\
    \Xhline{2\arrayrulewidth}
    \multicolumn{4}{l}{\textbf{(b) Student Model}} \\
    DREEAM & \textbf{63.47}\textsubscript{\textpm{0.02}} & 65.30\textsubscript{\textpm{0.03}} & \textbf{55.68}\textsubscript{\textpm{0.04}} \\
    \textit{w/o} ER self-training & 61.96\textsubscript{\textpm{0.39}} & 63.77\textsubscript{\textpm{0.44}} & 53.72\textsubscript{\textpm{0.43}}\\
    \textit{w/o} ER fine-tuning & 63.34\textsubscript{\textpm{0.02}} & \textbf{65.50}\textsubscript{\textpm{0.02}} & 55.27\textsubscript{\textpm{0.05}}\\
    \textit{w/o} both & 62.13\textsubscript{\textpm{0.07}} & 63.82\textsubscript{\textpm{0.08}} & 47.13\textsubscript{\textpm{0.12}} \\
    \Xhline{3\arrayrulewidth}
    \end{tabular}
    \caption{Ablation studies evaluated on the DocRED development set.}
    \label{tab:ablation}
\end{table}

This subsection investigates the effect of evidence-guided attention by ablation studies.
All subsequent experiments adopt BERT\textsubscript{base} as the PLM encoder.
We report scores without the inference-stage fusion strategy~\cite{xie-etal-2022-eider}.

\paragraph{Teacher Model} Firstly, we examine how guiding attention with evidence helps RE training on human-annotated data.
We train a variant of our teacher model without ER training and evaluate its performance on the DocRED development set.
In general, disabling ER training reduces the model to a baseline similar to ATLOP~\cite{zhou2021atlop}\footnote{The difference between ATLOP and our baseline is that our baseline utilizes the last three layers of PLM to obtain embeddings, whereas ATLOP adopts only the final layer.}. 

As presented in Table~\ref{tab:ablation} (a), the RE performance of our system decreases without ER training.
This observation supports the hypothesis that guiding attention with evidence is beneficial to improving RE.
We further visualize the token importance $\bm{q}^{(s,o)}$ for some instances to investigate the effect of evidence-guided training and find that our method succeeds in guiding the attention to focus more on relevant contexts.
The details can be found in Appendix~\ref{sec:visualization}.

Additionally, we retrieve evidence from the ER-disabled model as sentences with importance higher than the pre-defined threshold.
By doing so, we find that the Evi F1 is not far from its evidence-aware counterpart.
This observation indicates that ER is a task highly coupled with RE.

\paragraph{Student Model} Next, we investigate the student model trained on distantly-supervised data and finetuned on human-annotated data.
The aim is to examine the effect of guiding attention with evidence at various stages of training.
To this end, we remove ER supervisory signals from the student model during the training on distantly-supervised and human-annotated data.
The baseline excludes ER supervision from both stages, pre-trained on distantly-supervised data and then finetuned on human-annotated data for only RE.

As shown in Table~\ref{tab:ablation} (b), DREEAM without ER self-training performs comparably to the baseline, while DREEAM without ER fine-tuning performs comparably to the original model with no ablations.
These results indicate that ER self-training is more essential than ER fine-tuning for the student model.
On the one hand, we observe that disabling ER self-training on massive data causes a huge loss of evidence knowledge that cannot be recovered by finetuning on the much smaller evidence-annotated dataset.
On the other hand, we can conclude that DREEAM succeeds in retrieving evidence knowledge from the data without any evidence annotation, demonstrating the effectiveness of our ER self-training strategy.


\subsection{Memory Efficiency}

\label{sec:memory}

\begin{table}[t]
    \centering
    \small
    \begin{tabular}{lcc}
    \Xhline{3\arrayrulewidth}
    Method &  Memory & Trainable \\
    & (GiB) & Params. (M) \\
     \Xhline{3\arrayrulewidth}
    \multicolumn{3}{l}{\textbf{(a) without ER Module}} \\
    ATLOP~\cite{zhou2021atlop} & 10.8 & 115.4\\
    SSAN~\cite{xu-etal-2021-ssan} & 6.9 & 113.5\\
    KD-DocRE~\cite{tan-etal-2022-document} & 15.2 & 200.1 \\
    \Xhline{2\arrayrulewidth}
    \multicolumn{2}{l}{\textbf{(b) with ER Module}} \\
    EIDER~\cite{xie-etal-2022-eider} & 43.1 & 120.2 \\
    SAIS~\cite{xiao-etal-2022-sais} & 46.2 & 118.0 \\
    DREEAM (proposed) & 11.8 & 115.4 \\
    \Xhline{3\arrayrulewidth}
    \end{tabular}
    \caption{Memory consumption and the number of trainable parameters of DREEAM and existing methods.}
    \label{tab:memory}
\end{table}

This subsection discusses the memory inefficiency issue in previous ER approaches and shows how DREEAM solves it.
Previous approaches regard ER as a separate task from RE that requires extra neural network layers to solve~\cite{huang-etal-2021-entity,xie-etal-2022-eider,xiao-etal-2022-sais}.
To perform ER, all of them introduce a bilinear evidence classifier that receives an entity-pair-specific embedding and a sentence embedding as inputs.
For example, EIDER computes an evidence score for sentence $x_i$ with regard to entity pair $(e_s,e_o)$ as below:
\begin{equation}
    \mathrm{P}(x_i|e_s,e_o) = \sigma (\bm{x}_i\mathsf{W}\bm{c}^{(s,o)} + \bm{b}),
    \label{eq:eider}
\end{equation}
where $\bm{x}_i$ is a sentence embedding, $\bm{c}^{(s,o)}$ is the localized context embedding computed from Equation~\ref{eq:context_rep}, $\mathsf{W}$ and $\bm{b}$ are trainable parameters.
EIDER and other existing systems thus need to compute over all combinations of (sentence, entity pair).
Specifically, consider a document $D$ with $n$ sentences $\setX_D=\{x_1,x_2,\dots,x_n\}$ and $m$ entities $\setE_D=\{e_1,e_2,\dots,e_m\}$, yielding $m\times(m-1)$ entity pairs.
To obtain evidence scores, EIDER must perform bilinear classification $n\times m\times(m-1)$ times via Equation~\ref{eq:eider}, resulting in huge memory consumption.
In contrast, DREEAM takes the summations of attention weights over tokens as evidence scores, thus introducing neither new trainable parameters nor expensive matrix computations.
Hence, we see that DREEAM is more memory-efficient than its competitors.

Table~\ref{tab:memory} summarizes the memory consumption and the number of trainable parameters when utilizing BERT\textsubscript{base} as the PLM encoder for existing and proposed methods.
Values are measured when training the systems using the corresponding official repositories with a batch size of four\footnote{The value of EIDER is different from the original paper because we enable ER evaluations during training.}.
We observe that the memory consumption of DREEAM is only 27.4\% of EIDER and 25.5\% of SAIS.
Notably, DREEAM also consumes less memory than KD-DocRE, underscoring the memory efficiency of our proposed method. 

\subsection{Performance on Re-DocRED}

\begin{table}[!t]
    \centering
    \small
    \begin{tabular}{lrr}
    \Xhline{3\arrayrulewidth}
    \multicolumn{1}{c}{\textbf{Statistics}} & \multicolumn{1}{c}{DocRED} & \multicolumn{1}{c}{Re-DocRED} \\
    \Xhline{2\arrayrulewidth}
    \# rel. & 38,180 & 85,932 \\
    \# rel. w/o evi. & 1,421 (3.7\%) & 38,672 (45.0\%)\\
    \Xhline{3\arrayrulewidth}
    \end{tabular}
    \caption{Statistics of relation triples in the training set of DocRED and Re-DocRED. \textit{rel.} stands for relation triples and \textit{rel. w/o evi.} stands for relation triples without evidence sentences.}
    \label{tab:evi_redocred}
\end{table}

\begin{table}[!t]
    \small
    \centering
    \begin{tabular}{lcc}
    \Xhline{3\arrayrulewidth}
         Method & Ign F1 & F1 \\ 
    \Xhline{2\arrayrulewidth}
    \multicolumn{3}{l}{\textbf{(a) without Distantly-Supervised Data}}\\
    ATLOP~\cite{zhou2021atlop}   & 76.82 & 77.56 \\
    DocuNet~\cite{zhang-etal-2021-document} & 77.26 & 77.87 \\
    KD-DocRE~\cite{tan-etal-2022-document}  & 77.60 & 78.28 \\
    \hdashline
    DREEAM & 77.34\textsubscript{\textpm{0.19}} & 77.94\textsubscript{\textsubscript{\textpm{0.15}}} \\
    + Inference-Stage Fusion & 79.66\textsubscript{\textpm{0.39}} & 80.73\textsubscript{\textpm{0.38}} \\
    \Xhline{2\arrayrulewidth}
    \multicolumn{3}{l}{\textbf{(b) with Distantly-Supervised Data}} \\
    ATLOP~\cite{zhou2021atlop}  & 78.52 & 79.46 \\
    DocuNet~\cite{zhang-etal-2021-document}  & 78.52 & 79.46 \\
    KD-DocRE~\cite{tan-etal-2022-document}  & 80.32 & 81.04 \\
    \hdashline
    DREEAM & 78.67\textsubscript{\textpm{0.17}} & 79.35\textsubscript{\textpm{0.18}} \\
    +Inference-Stage Fusion & \textbf{80.39}\textsubscript{\textpm{0.03}} & \textbf{81.44}\textsubscript{\textpm{0.04}} \\
    \Xhline{3\arrayrulewidth}
    \end{tabular}
    \caption{Evaluation results on the test set of Re-DocRED, with best scores \textbf{bolded}. PLM encoder is aligned to RoBERTa-large. The scores of existing methods are borrowed from~\citet{redocred}.}
    \label{tab:re-docred}
\end{table}

Although DocRED is a widely used benchmark, recent works have pointed out that annotations of the dataset are incomplete~\cite{huang-etal-2022-recommend,xie-etal-2022-eider,redocred}.
To paraphrase, many relation triples in DocRED are missing in human annotations, biasing the dataset with many false negatives.
\citet{redocred} thus proposed Re-DocRED, a more reliable benchmark for DocRE that revises DocRED to alleviate the false negative issue.
In this subsection, we evaluate DREEAM on Re-DocRED to verify the soundness of our proposed method.

Similar to Section~\ref{sec:main_results}, we conducted experiments under two different settings:
(a) a fully-supervised setting without distantly-supervised data and (b) a weakly-supervised setting utilizing distantly-supervised data.
Notably, Re-DocRED introduces new relation triples without providing accurate evidence sentences.
As shown in Table~\ref{tab:evi_redocred}, compared with DocRED, the training set of Re-DocRED contains much more relation triples without evidence sentences.
DREEAM trained on Re-DocRED could thus be inaccurate on ER, biased by the considerable amount of missing evidence.
Therefore, during ER self-training of the student model, we adopt token evidence distributions predicted by a teacher model trained on DocRED as the supervisory signal.
The student model is further finetuned on Re-DocRED to obtain more reliable knowledge about RE. 

Table~\ref{tab:re-docred} compares the performance of DREEAM against existing methods.
We observe that DREEAM outperforms existing methods under both the fully-supervised setting and the weakly-supervised setting.
The observation indicates the soundness of our proposed method.

\section{Related Work}

\paragraph{DocRE} Recent work has extended the scope of relation extraction task from sentence to document~\cite{peng-etal-2017-cross,quirk-poon-2017-distant,yao-etal-2019-docred}. 
Compared with its sentence-level counterpart, DocRE is a more realistic and challenging setting, aiming at extracting both intra- sentence and inter-sentence relations.
Although commonly-used benchmarks for DocRE include DocRED~\cite{yao-etal-2021-adapt}, CDR~\cite{li2016CDR} and GDA~\cite{wu2019GDA}, only DocRED contains evidence annotation and massive pre-processed data obtained from relation distant supervision.
Therefore, we adopt DocRED as our test bed.

\paragraph{Transformer-based DocRE} Modeling DocRE with a Transformer-based system has been a popular and promising approach, outperforming its graph-based counterparts~\cite{zeng-etal-2020-double,zeng-etal-2021-sire,xu-etal-2021-reconstruct}.
One of the major topics of these systems is a better utilization of long-distance token dependencies captured by the PLM encoder.
\citet{zhang-etal-2021-document} formulate DocRE as a semantic segmentation task and introduce a U-Net~\cite{ronneberger-etal-2015-unet} on top of the PLM encoder to capture local and global dependencies between entities.
\citet{zhou2021atlop} propose localized contextual pooling to focus on tokens relevant to each entity pair.
Based on their work, \citet{tan-etal-2022-document} adopt an axial attention module to perform two-hop reasoning and capture the dependencies between relation triples.
These designs provide no supervision on token dependencies, expecting the model to capture them implicitly during training.
In contrast, we provide explicit supervision for token dependencies by utilizing evidence information.

\paragraph{ER in DocRE} This study is not the first to incorporate evidence information into DocRE.
\citet{huang-etal-2021-three} first report that heuristically selecting evidence sentences boosts the performance of DocRE systems.  
\citet{huang-etal-2021-entity}, \citet{xie-etal-2022-eider} and \citet{xiao-etal-2022-sais} train neural classifiers to automatically retrieve evidence together with RE.
However, we perform ER with neither heuristic rules nor neural classifiers.
Furthermore, our approach can be used for ER self-training on data without evidence annotations.

\paragraph{Distant Supervision} Distant supervision has been widely adopted as a technique to generate automatically-labeled data for RE~\cite{mintz-etal-2009-distant,quirk-poon-2017-distant,xiao-etal-2020-denoising}.
The method assumes that if a sentence contains an entity pair that participates in a known relation in a knowledge base (KB), the sentence probably expresses that relation.
Thus unlabeled text can be aligned with a KB using entities as anchors, with each match distantly supervised by the relation described in the KB.
\citet{yao-etal-2019-docred} apply the technique to annotate relations in documents automatically.
In this work, we directly adopt those documents for ER self-training.

\section{Conclusion}
We have introduced methods to improve the usage of ER in DocRE.
First, we propose DREEAM, a memory-efficient method to reduce the computation cost of ER.
Unlike existing approaches that train an evidence classifier for ER, DREEAM directly supervises the attention to concentrate more on evidence than on others.
Next, we propose to employ DREEAM in a weakly-supervised setting to compensate for the shortage of human annotations.
Instead of gold evidence annotated by humans, we adopt evidence predictions from a teacher model trained on human-annotated data as the supervisory signal to realize ER self-training on unlabeled data.
Experiments on the DocRED benchmark show that DREEAM exhibits state-of-the-art performance on both RE and ER, with the help of weakly-supervised training on data obtained from distant supervision of relations. 
Compared with existing approaches, DREEAM performs ER with zero trainable parameters introduced, thereby reducing the memory usage to 27\% or less.
The soundness of DREEAM is confirmed by conducting experiments on Re-DocRED, a revised version of DocRED.

In the future, we plan to transfer the evidence knowledge of DREEAM trained on DocRED to other DocRE datasets. 

\section*{Limitations}
A major limitation of this work is that our method can only retrieve relation-agnostic evidence.
Unlike~\citet{xiao-etal-2022-sais}, DREEAM cannot specify evidence sentences for each relation label.
Therefore, when an entity pair holds multiple relations, DREEAM retrieves the same evidence regardless of the relation type, even though the evidence may be correct for some of the relations but not for others.

\section*{Ethics Statement}
In this work, we have proposed a method for incorporating ER into DocRE.
Our approach directly supervises the weights of attention modules within a Transformer-based PLM encoder. Inside the research community, we hope our approach can provide a new viewpoint on the explainability of document-level relation extraction systems.
Furthermore, a better DocRE system will benefit the research on other tasks, such as question answering and reading comprehension.
In the real world, a DocRE system with good performance can help extract useful information from unstructured text, reducing human efforts and expenses. 
Furthermore, as our method is memory-efficient, it is also friendly to the environment.

We also have demonstrated a use case of our method in ER self-training, utilizing massive data obtained from relation distant-supervision.
Although in this work, we directly adopt the data provided by~\citet{yao-etal-2019-docred}, it is possible to extend the scale of data by utilizing numerous unstructured texts.
Utilizing a wide range of unstructured texts may expose our system to the risk of vulnerable data, potentially biasing our system in the wrong direction.
To mitigate the problem, we encourage performing data pre-processing to detect and remove harmful contents before training.

\section*{Acknowledgements}

This paper is based on results obtained from a project, JPNP18002, commissioned by the New Energy and Industrial Technology Development Organization (NEDO).
We appreciate Marco Cognetta for the generous help and valuable discussions.

\bibliography{anthology,custom}
\bibliographystyle{acl_natbib}

\appendix

\section{Hyper-Parameters and Runtime}
\label{sec:hyparam}

We adopt AdamW as the optimizer~\cite{loshchilov2018decoupled} and apply a linear warmup for the learning rate at the first 6\% steps.
Important hyper-parameters are shown in Table~\ref{tab:hyparam}, which are mainly borrowed from existing works.
Specifically, we borrow hyper-parameters from~\citet{zhou2021atlop} to train the teacher model and borrow those from~\citet{tan-etal-2022-document} to train and finetune the student model. 
The only exception is the number of epochs for training the student model, which is determined by a grid search from $\{2, 5, 8, 10\}$.

The average running time spent for our system at each training stage is shown in Table~\ref{tab:time}.
Note that we employ a single Tesla V100 16GB GPU when utilizing BERT\textsubscript{base} and a single NVIDIA A100 40GB GPU when utilizing RoBERTa\textsubscript{large}.

\begin{table*}[t]
    \centering
    \small
    \begin{tabular}{lcccccc}
    \Xhline{3\arrayrulewidth}
    Hyperparam. & \multicolumn{2}{c}{Train (teacher)} & \multicolumn{2}{c}{Train (student)} & \multicolumn{2}{c}{Finetune (student)} \\
    & BERT\textsubscript{base} & RoBERTa\textsubscript{large} & BERT\textsubscript{base} & RoBERTa\textsubscript{large} & BERT\textsubscript{base} & RoBERTa\textsubscript{large} \\
    \Xhline{3\arrayrulewidth}
         \# Epoch &  30 & 30 & 2 & 5 & 10 & 10 \\
         lr for encoder & 5e-5 & 3e-5 & 3e-5 & 1e-5 & 1e-6 & 1e-6 \\
         lr for classifier & 1e-4 & 1e-4 & 1e-4 & 5e-5 & 3e-6 & 3e-6\\
         max gradient norm & 1.0 & 1.0 & 5.0 & 5.0 & 2.0 & 2.0 \\
    \Xhline{3\arrayrulewidth}
    \end{tabular}
    \caption{Hyper-parameters in training.}
    \label{tab:hyparam}
\end{table*}

\begin{table}[t]
    \centering
    \small
    \begin{tabular}{lcc}
    \Xhline{3\arrayrulewidth}
     Phase & BERT\textsubscript{base} & RoBERTa\textsubscript{large} \\
    \Xhline{3\arrayrulewidth}
    Train (teacher) & 1h18min & 1h18min \\
    Train (student) & 2h55min & 6h12min \\
    Finetune (student) & 26min & 29min \\
    \Xhline{3\arrayrulewidth}
    \end{tabular}
    \caption{Runtime for each training stage.}
    \label{tab:time}
\end{table}

\section{Visualization: Evidence-Guided Attention}
\label{sec:visualization}

\begin{figure}[t!]
    \centering
    \subfigure[Before attention guidance.]{\includegraphics[width=.5\textwidth]{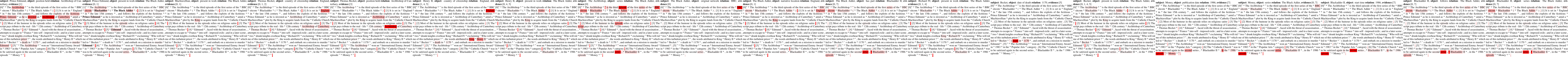}}
    \subfigure[After attention guidance.]{\includegraphics[width=.5\textwidth]{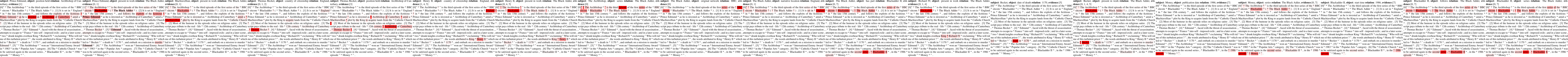}}
    \caption{Heatmaps of token importance for localized context pooling before and after guiding the attention with evidence when deciding the relation for entity pair (\textit{Prince Edmund}, \textit{The Black Adder}). The gold relation is \textit{present in work} with evidence sentences 1 and 2. Deeper the color, the larger the value. }
    \label{fig:heatmap_1}
\end{figure}

\begin{figure}[t!]
    \centering
    \subfigure[Before attention guidance.]{\includegraphics[width=.5\textwidth]{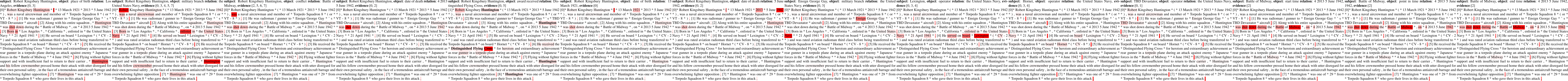}}
    \subfigure[After attention guidance.]{\includegraphics[width=.5\textwidth]{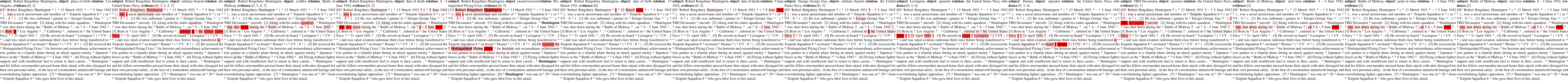}}
    \caption{Heatmaps of token importance for localized context pooling before and after guiding the attention with evidence when deciding the relation for entity pair (\textit{Robert Kingsbury Huntington}, \textit{Distinguished Flying Cross}). The gold relation is \textit{award received} with evidence sentences 1 and 6. Deeper the color, the larger the value. }
    \label{fig:heatmap_2}
\end{figure}

As introduced in Section~\ref{sec:teacher}, evidence knowledge of DREEAM originates from sentence-level supervision.
We hypothesize that sentence-level supervision, from a more macro perspective, should improve its micro counterpart of token-level focusing.
To test the hypothesis, we examine the token-level evidence distribution for localized context pooling.
Specifically, we utilize heatmaps to visualize $\bm{q}^{(s,o)}$ and observe the differences before and after evidence-guided training.

Results are shown in Figure~\ref{fig:heatmap_1} and ~\ref{fig:heatmap_2}. 
We adopt the toolkit developed by~\citet{yang-zhang-2018-ncrf}.
It is obvious that the distribution is more focused on sentences 1 and 2 in Figure~\ref{fig:heatmap_1}(b) than in Figure~\ref{fig:heatmap_1}(a).
Before training the evidence-guided attention, the model tends to focus on the period of each sentence.
Guiding the attention with evidence helps the model to focus more on sentences 1 and 2, as well as the critical tokens providing a clue for relation classification, such as \textit{fictitious} in Figure~\ref{fig:heatmap_1} and \textit{received} in Figure~\ref{fig:heatmap_2}.

\end{document}